\documentclass[letterpaper]{article} %
\usepackage{aaai2026}  %
\usepackage{times}  %
\usepackage{helvet}  %
\usepackage{courier}  %
\usepackage[hyphens]{url}  %
\usepackage{graphicx} %

\usepackage{amsmath}
\usepackage{booktabs}       %
\usepackage{multirow}       %
\usepackage{adjustbox}      %
\usepackage{graphicx}       %
\usepackage{caption}        %

\usepackage{natbib}  %
\usepackage{caption} %
\usepackage{algorithm}
\usepackage{algorithmic}

\usepackage{newfloat}
\usepackage{listings}
\DeclareCaptionStyle{ruled}{labelfont=normalfont,labelsep=colon,strut=off} %
\floatstyle{ruled}
\newfloat{listing}{tb}{lst}{}
\floatname{listing}{Listing}
\title{MM-R1: Unleashing the Power of Unified Multimodal Large Language Models for Personalized Image Generation}
\author{
    Qian Liang\textsuperscript{\rm 1}\thanks{These authors contributed equally.}\\ Yujia Wu\textsuperscript{\rm 1}\footnotemark[1], Kuncheng Li\textsuperscript{\rm 1}, Jiwei Wei\textsuperscript{\rm 1}\thanks{Corresponding author.}, Shiyuan He\textsuperscript{\rm 1}, Jinyu Guo\textsuperscript{\rm 1}, Ning Xie\textsuperscript{\rm 1}
}
\affiliations{
    \textsuperscript{\rm 1}University of Electronic Science and Technology of China\\

}

\usepackage{bibentry}

\nocopyright

\begin{document}

\maketitle

\begin{abstract}
Multimodal Large Language Models (MLLMs) with unified architectures excel across a wide range of vision-language tasks, yet aligning them with personalized image generation remains a significant challenge. Existing methods for MLLMs are frequently subject-specific, demanding a data-intensive fine-tuning process for every new subject, which limits their scalability. In this paper, we introduce \textbf{MM-R1}, a framework that integrates a cross-modal Chain-of-Thought (X-CoT) reasoning strategy to unlock the inherent potential of unified MLLMs for personalized image generation. Specifically, we structure personalization as an integrated visual reasoning and generation process: (1) grounding subject concepts by interpreting and understanding user-provided images and contextual cues, and (2) generating personalized images conditioned on both the extracted subject representations and user prompts. To further enhance the reasoning capability, we adopt Grouped Reward Proximal Policy Optimization (GRPO) to explicitly align the generation. Experiments demonstrate that MM-R1 unleashes the personalization capability of unified MLLMs to generate images with high subject fidelity and strong text alignment in a zero-shot manner.
\end{abstract}

\section{Introduction}

With the rapid advancement of Multimodal Large Language Models (MLLMs), the capabilities of unified architectures for vision–language understanding and generation have increasingly converged and mutually reinforced one another. Recent advances \cite{Chameleon_Team_Chameleon_Mixed-Modal_Early-Fusion_2024, liu2024lumina, xie2025show, zou2025omnimambaefficientunifiedmultimodal} have shown that these architectures can effectively handle a variety of vision-language tasks, such as visual question answering and text-to-image generation, demonstrating strong multimodal understanding and generation capabilities. However, their potential in more fine-grained tasks, particularly personalized image synthesis, remains insufficiently explored.
\begin{figure}[t]
  \centering
  \includegraphics[width=\linewidth]{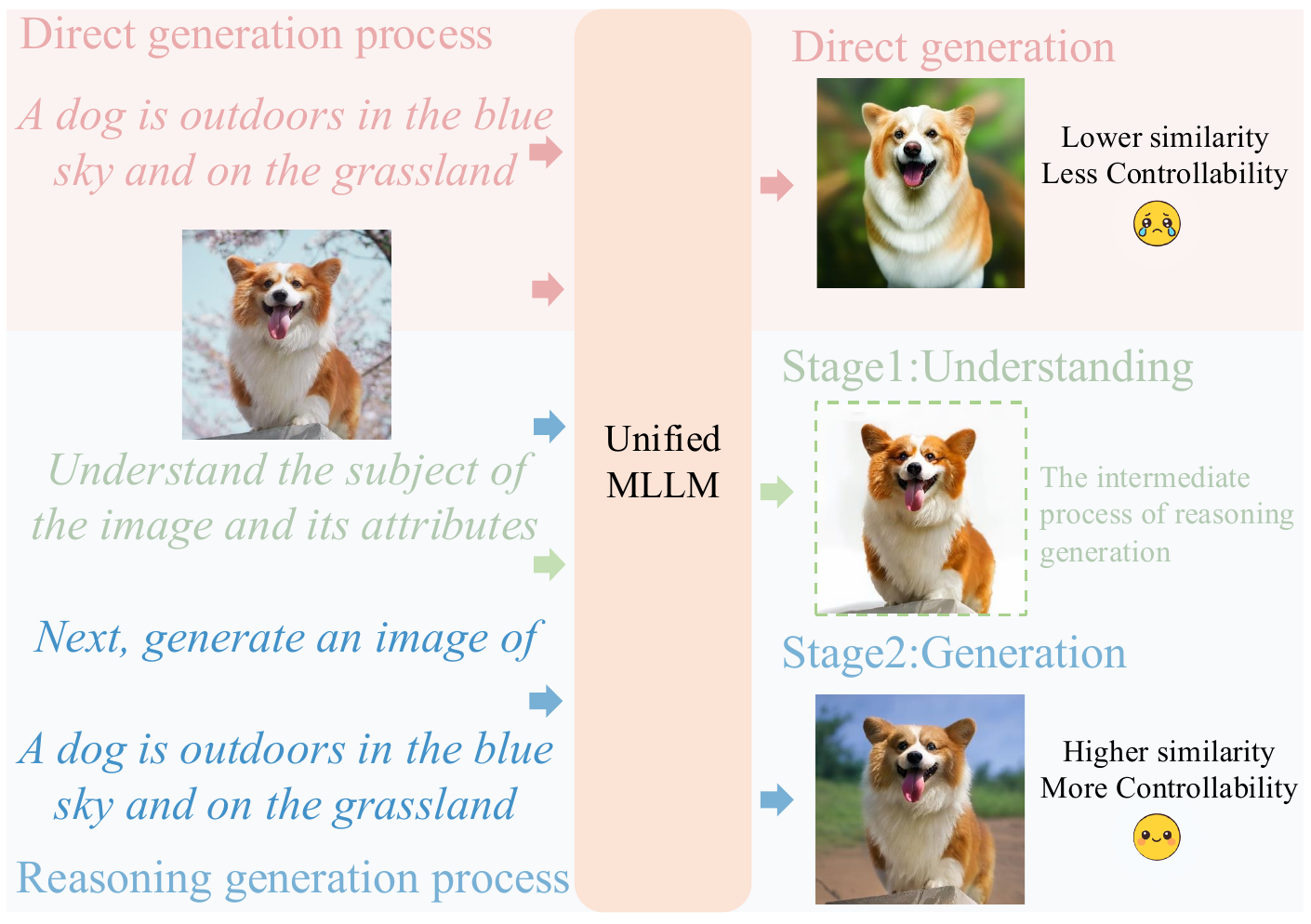}
  \caption{\textbf{Comparison between reasoning generation and ordinary generation.} Reasoning generation first understands the subject and its attributes in the image, and then injects these attributes into the generation process.}
  \label{fig:intro_image}
\end{figure}

Existing studies on personalizing unified MLLMs for image generation often involve training subject-specific tokens using the specific dataset. For instance, \cite{sun2025personalized} performed subject-specific fine-tuning by creating trainable parameters for each subject. Yo'chameleon \cite{nguyen2025yo} introduces soft prompts as learnable tokens to represent user concepts during both understanding and generation. UniCTokens \cite{an2025unictokens} proposed unified concept tokens refined through progressive staged optimization. Despite methodological differences, these approaches all rely on external token mechanisms and subject-specific optimization, which limit the scalability and generalization of personalization heavily.

Recently, some works \cite{sun2024generative, sun2024x, wu2025proxy} established personalization as a two-fold task: it first requires the model to comprehend the user-specified subject, and then to generate images that are faithful to both this subject and the accompanying prompt. This principle, the tight coupling of understanding and generation, mirrors the core design philosophy of unified MLLMs, which are engineered to integrate the understanding and generating within a single, coherent framework \cite{an2025unictokens}. This inherent alignment suggests a promising yet underexplored path: achieving personalization by directly enhancing the model's intrinsic reasoning abilities, unlocking significant potential for future development and applications. 

To validate this, we conducted a preliminary trial of a two-step personalization process as illustrated in Fig. \ref{fig:intro_image}. By prompting an MLLM \cite{liu2024lumina} to first analyze the attributes of a subject in an image and then generate, it synthesizes significantly better results even without any additional training. This observation supports for our hypothesis and logically leads to our proposed solution: a unified framework that achieves personalization by deeply integrating reasoning.

In this paper, we propose \textbf{MM-R1}, a reasoning-enhanced framework for personalized image synthesis based on unified MLLMs. MM-R1 integrates a cross-modal reasoning mechanism with reward-guided optimization to better align the model’s understanding and generation capabilities with user-defined content. Inspired by how Chain-of-Thought (CoT) improves reasoning in LLMs \cite{kojima2022large, wei2022emergent, wang2022self}, we begin by introducing a fundamental cross-modal Chain-of-Thought (\textbf{X-CoT}) strategy, which explicitly defines the personalization as visual understanding and generation process. In the understanding stage, the model deconstructs the user-provided image and contextual cues to distill two key outputs: an explicit textual description of the subject, and an intermediate ``focus image'' that visually isolates the identified concept. Subsequently, in the generation stage, the model synthesizes the final scene by seamlessly weaving together its understanding of these extracted representations with the user-specified prompt, thereby formulating a coherent layout and rendering a personalized image faithful to both the subject's identity and the prompt's semantics. %

To this, we design an X-CoT Data Engine. This automated pipeline generates high-quality, cross-modal Chain-of-Thought annotations that mirror our desired reasoning path. Thus, the model receives structured supervision, learning to explicitly link the initial subject concept with the final conditioned generation. This foundational training helps learn the core reasoning patterns necessary for enhancing subject fidelity and text alignment. Considering X-CoT is used as cold-start, we further employ Reinforcement Learning (RL). Specifically, we adopt the Group Relative Policy Optimization (GRPO) strategy \cite{deepseekai2025deepseekr1incentivizingreasoningcapability}. GRPO enables fine-grained, reward-guided optimization by using multifaceted signals, such as image similarity for subject fidelity and text-image consistency for prompt adherence, to directly evaluate and enhance the generated outputs. 
In essence, our approach combines supervised pre-training with reinforcement learning fine-tuning, creating a principle-based, end-to-end framework that fully leverages the reasoning and generation capabilities of a unified MLLM for effective and controllable personalized image synthesis.

In summary, our main contributions are as follows:
\begin{itemize}
    \item We propose MM-R1, a framework that equips unified MLLMs with personalized image synthesis capabilities through a cross-modal reasoning strategy (X-CoT) combining subject grounding and conditioned generation.
    \item We develop an X-CoT Data Engine, an automated pipeline that generates structured cross-modal reasoning annotations for models to complete personalized tasks, thereby achieving training without manual labeling.
    \item We extend GRPO to personalized image synthesis with multi-aspect rewards, improving subject fidelity, prompt alignment, and controllability.
    \item Our experiments demonstrate strong zero-shot personalization and superior performance over existing methods in both fidelity and controllability.
\end{itemize}

\section{Related Work}
\subsection{Personalized Image Synthesis}
Personalized image synthesis focuses on generating images of specific subjects in diverse contexts while maintaining subject fidelity and prompt alignment. Diffusion-based approaches have advanced this task substantially, with optimization-based methods such as DreamBooth \cite{ruiz2023dreambooth}, Custom Diffusion \cite{kumari2023multi}, and Textual Inversion \cite{gal2022image} fine-tuning model weights or token embeddings for each subject, achieving high fidelity but requiring costly per-subject optimization. To improve efficiency, tuning-free diffusion methods extract subject features and understand subject attributes with pre-trained encoders, enabling zero-shot or few-shot generation and inspiring more scalable personalization techniques~\cite{ye2023ip, labs2025flux}. More recently, unified MLLMs have been explored for personalized image synthesis, including autoregressive models \cite{Chameleon_Team_Chameleon_Mixed-Modal_Early-Fusion_2024} that adopt a two-stage training strategy combining text embedding optimization and transformer fine-tuning, as well as methods such as Yo'Chameleon \cite{nguyen2025yo} and UniCTokens \cite{an2025unictokens}, which inject subject information into unified MLLMs via soft prompts or concept tokens to support both understanding and generation. Despite these advances, existing approaches still rely on separate per-subject representations, increasing complexity and limiting generalization. In contrast, our work focuses on personalized image synthesis within unified MLLMs, achieving efficient zero-shot personalization without disjoint subject-specific components or an additional encoder for understanding.

\subsection{Reasoning in Multimodal Large Language Models}

\begin{figure*}[htbp]
   \centering
   \includegraphics[width=\linewidth]{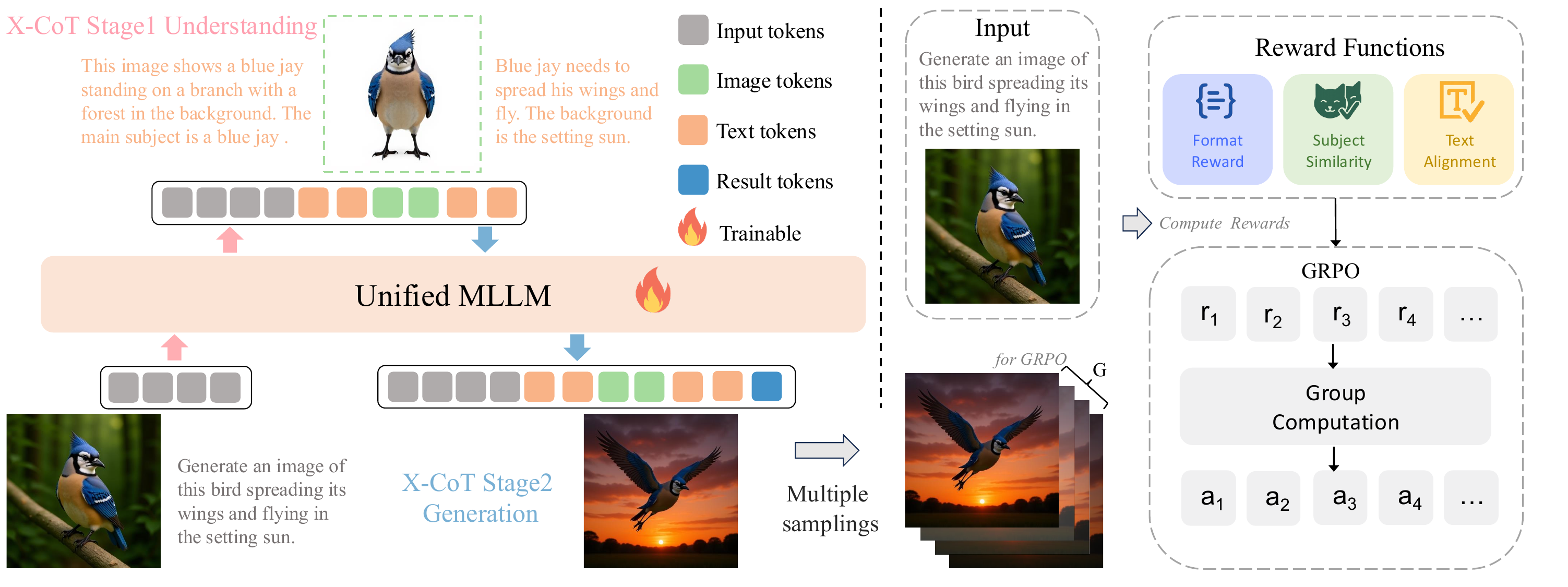}
   \caption{\textbf{Schematic diagram of the method proposed in this paper.} The left part is our X-CoT process. The model first understands the user input to obtain the subject image and then generates the image. After that, for each sample, multiple outputs are generated and trained together with the user input for GRPO. In the reinforcement learning process, three reward functions are used to calculate the rewards for different generated results, and finally these rewards are used to adjust the model.
   }
   \label{fig:method_image}
\end{figure*}

Chain-of-Thought (CoT) reasoning \cite{wei2022chain,zhang2023automatic, kojima2022large} has emerged as a powerful mechanism to enhance the reasoning capabilities of LLMs by breaking complex problems into intermediate steps, improving both transparency and performance. Building on these advances, recent work has extended CoT to the multimodal setting, giving rise to MLLMs \cite{liu2023llava,team2024gemini, thawakar2025llamav} capable of processing both textual and visual inputs. For instance, T2I-R1 \cite{jiang2025t2i} and GoT-R1 \cite{duan2025got} respectively enable the model to generate images that better meet the user's requirements and those that better conform to the spatial positional relationship requirements through CoT. Multimodal CoT explores strategies such as grounding visual inputs into symbolic representations \cite{gao2025interleaved}, generating visualizations to accompany reasoning \cite{zhao2025cot}, or producing visual rationales through cropping and zooming \cite{wang2025multimodal}. Visual Planning \cite{xu2025visual} further investigates whether reasoning can emerge purely within the visual modality by structuring the reasoning process as a sequence of images without textual mediation. In contrast to the aforementioned approaches, our work explicitly incorporates cross-modal reasoning into unified MLLMs for personalized image synthesis, leveraging multi modalities in a unified and integrated manner that aligns with the pretraining paradigm and the foundational design principles of unified MLLMs.

\section{Method}
We present the MM-R1 framework, which enables unified MLLMs (e.g., Chameleon \cite{Chameleon_Team_Chameleon_Mixed-Modal_Early-Fusion_2024}, Lumina-mGPT~\cite{liu2024lumina}) to perform zero-shot personalized image generation by leveraging a novel cross-modal Chain-of-Thought (X-CoT) reasoning strategy and reinforcement learning methods GRPO. MM-R1 extends the capabilities of these models for personalized vision-language tasks, allowing them to generate personalized images without the need for subject-specific fine-tuning.

\subsection{Overview of MM-R1 Framework}

As depicted in Fig. \ref{fig:method_image}, our proposed MM-R1 framework is architected around a central cross-modal Chain-of-Thought (X-CoT) reasoning pipeline, which is fine-tuned using GRPO-based reinforcement learning. It enables zero-shot personalized image generation through a structured, two-step inference process.

First, in the understanding and planning phase, the X-CoT strategy processes multimodal inputs (including user images and text prompts) to generate a high-level ``generation blueprint''. This blueprint embodies a deep semantic understanding of the subject and a coherent plan for the final image generation. This initial step provides a robust, semantically-grounded foundation for the subsequent synthesis.
Next, in the generation phase, the model translates this abstract blueprint into a sequence of detailed visual tokens. This ensures the final rendered image is not only visually coherent but also strictly adheres to both the subject's identity and the prompt's semantics. 

Then, we use GRPO to teach the model to think better to improve the generation results. For each prompt, the model generates a batch of candidate images. These candidates then undergo a evaluation and ranking process based on a suite of predefined reward functions, which assess critical attributes like subject fidelity, text alignment, and format consistency. Ultimately, this process iteratively teaches the model to favor policies that produce images with higher subject consistency and stricter textual adherence.

\subsection{X-CoT Data Engine}
\begin{figure*}[htbp]
   \centering
   \includegraphics[width=\linewidth]{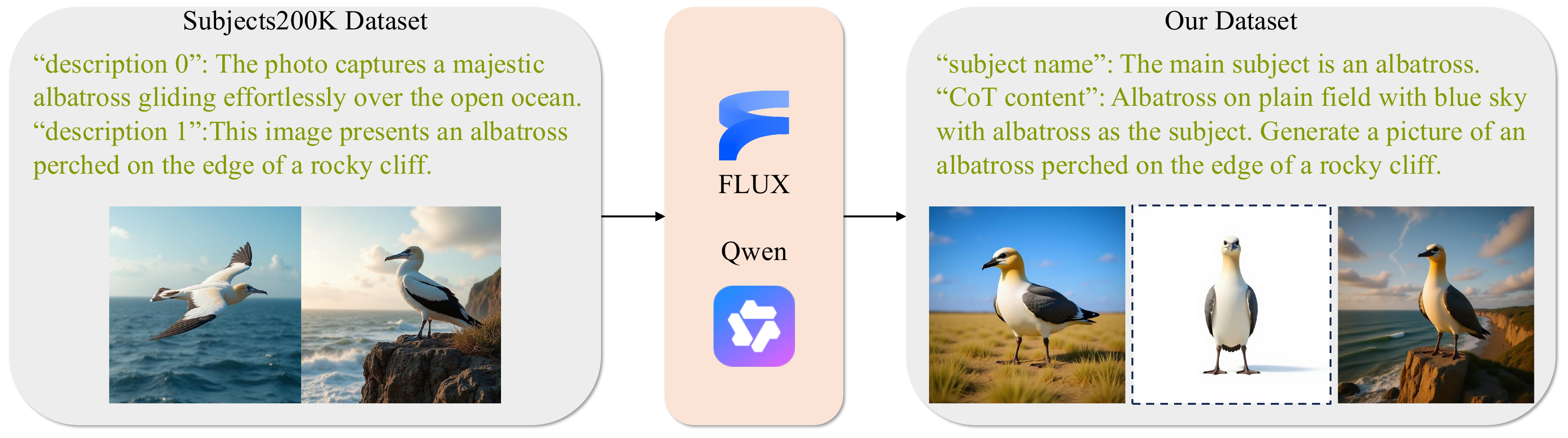}
   \caption{\textbf{An example of our data construction pipeline.} We reconstructed each sample in the Subjects200K dataset using FLUX-Kontext and Qwen2.5-VL-7B-Instruct respectively, and obtained three images and thinking content.
   }
   \label{fig:datapipeline_image}
\end{figure*}

Training our two-stage reasoning model requires a specialized dataset that goes beyond simple image-text pairs, but thinking content of the X-CoT. To this end, we engineered an automated data construction pipeline. We chose the Subjects200K dataset \cite{tan2025ominicontrol} as our foundation due to its rich diversity of subject categories. We reconstructed this dataset by regenerating images and texts using the FLUX-Kontext \cite{labs2025flux} and Qwen2.5-VL-7B-Instruct \cite{bai2025qwen2} model respectively. Fig. \ref{fig:datapipeline_image} shows an example creation process of our dataset.

Specifically, the Subjects200K dataset consists of samples from multiple categories, each of which contains two images with the same subject and the descriptive text of each image. We use FLUX-Kontext to first extract the subject image from each sample and then use this image and the description text of the dataset to generate the two images with the same subject we need while retaining the text.

Thinking contents consist of three parts. The first part is to understand the subject and its attributes of the reference image. We use the Qwen2.5-VL-7B-Instruct model to understand the reference image and generate the reference image title that meets the personalized generation task. The second part is the extraction of the subject by the model. Here, we directly use the subject image previously extracted. The third part of the text thinking contents is the integration and understanding of the reference image and the target text, which helps the model understand the paradigm of the target image. The Qwen2.5-VL-7B-Instruct model is also used to understand the content of the target image and obtain the corresponding generation prompt. 
We will present more details in the supplementary materials.

\subsection{Cold-Start Training with X-CoT}

By default, the model will directly integrate all the information and generate the result image when using unified MLLMs for personalized generation tasks. But this ability of the model is insufficient to achieve satisfactory generation results for personalized generation tasks \cite{wang2024emu3}. 
To address these issues, our approach is to have the model conduct reasoning before generating the results. The reasoning process includes understanding the content of the reference image and extracting the subject. To ensure the consistency of the final generated results with the reference images. During the reasoning process, we add a subject image to help the model better extract and understand the subject of the reference image, and at the same time the model uses the subject image rather than just user input to generate subsequent result images. The thinking contents include three parts, In the first part, the model conducts the necessary understanding of the reference image information. In the second part, for the personalized generation task, the model extracts the subject name and thinks obtaining the corresponding subject image. In the third part, based on the text prompt and all reference images, the model plans the content, layout, subject action, scene and other information of the resulting image. Finally, the model generates the result based on all the input information and the thinking content.

In order to enable the model to learn the ability to extract and utilize the subject information from the reference image through reasoning, we adopt a cold-start strategy based on supervised fine-tuning, dividing the generation process of the unified MLLMs into two stages. The first stage, the reasoning process, is used for the extraction and integration of information. The second stage utilizes this information to generate the resulting image. We utilized the constructed dataset to guide the model in generating reasoning contents in the expected format and semantics, strengthening the inherent information extraction and generation capabilities of the model to complete personalized generation tasks provides a stable and effective foundation for subsequent Reinforcement Learning.

\begin{figure*}[t]
   \centering
   \includegraphics[width=\linewidth]{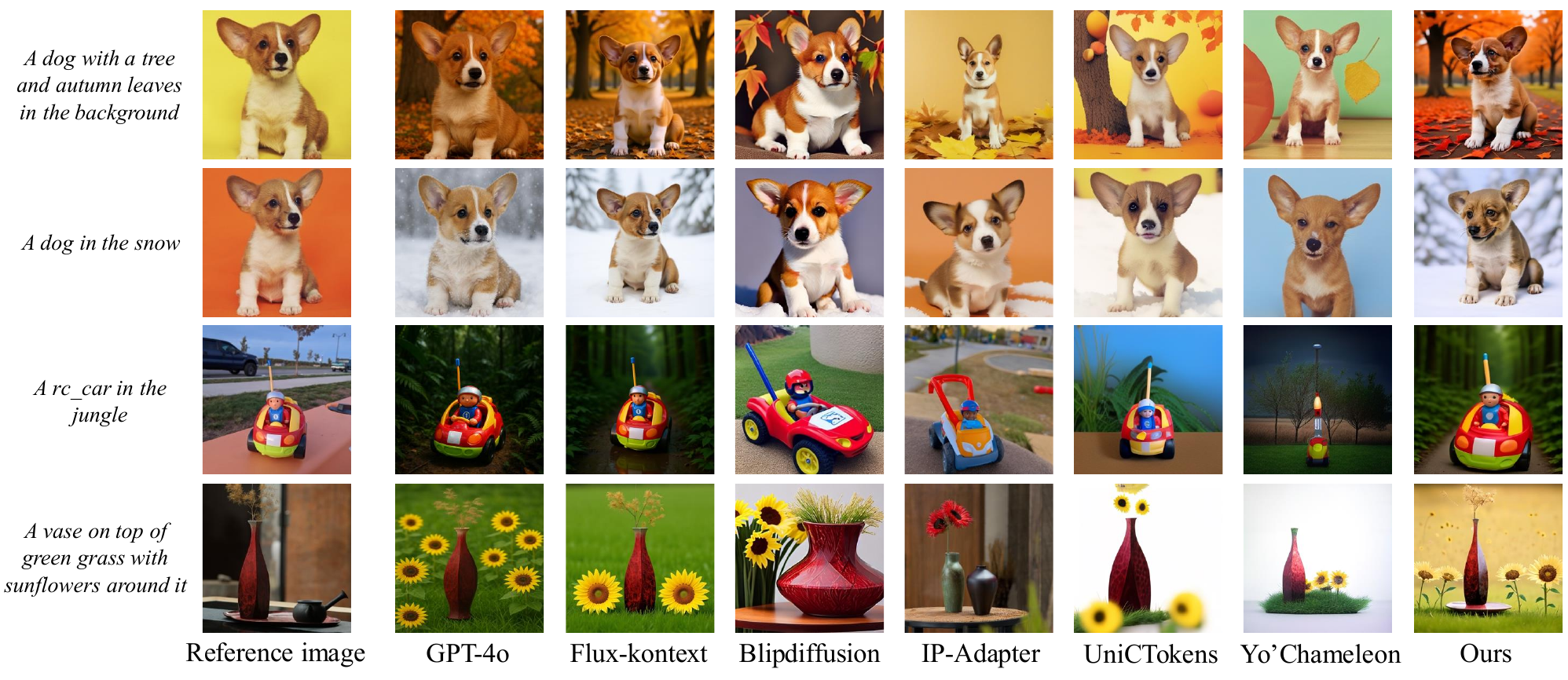}
   \caption{\textbf{Visualization of the qualitative results on DreamBench.} Our method can ensure extremely high subject similarity (the last column of the third row) and the aesthetic degree of the image (the last column of the fourth row). Meanwhile, our method can also generate diverse poses on the same subject.
   }
   \label{fig:duibijieguo_dbimage}
\end{figure*}

\begin{figure*}[h]
   \centering
   \includegraphics[width=\linewidth]{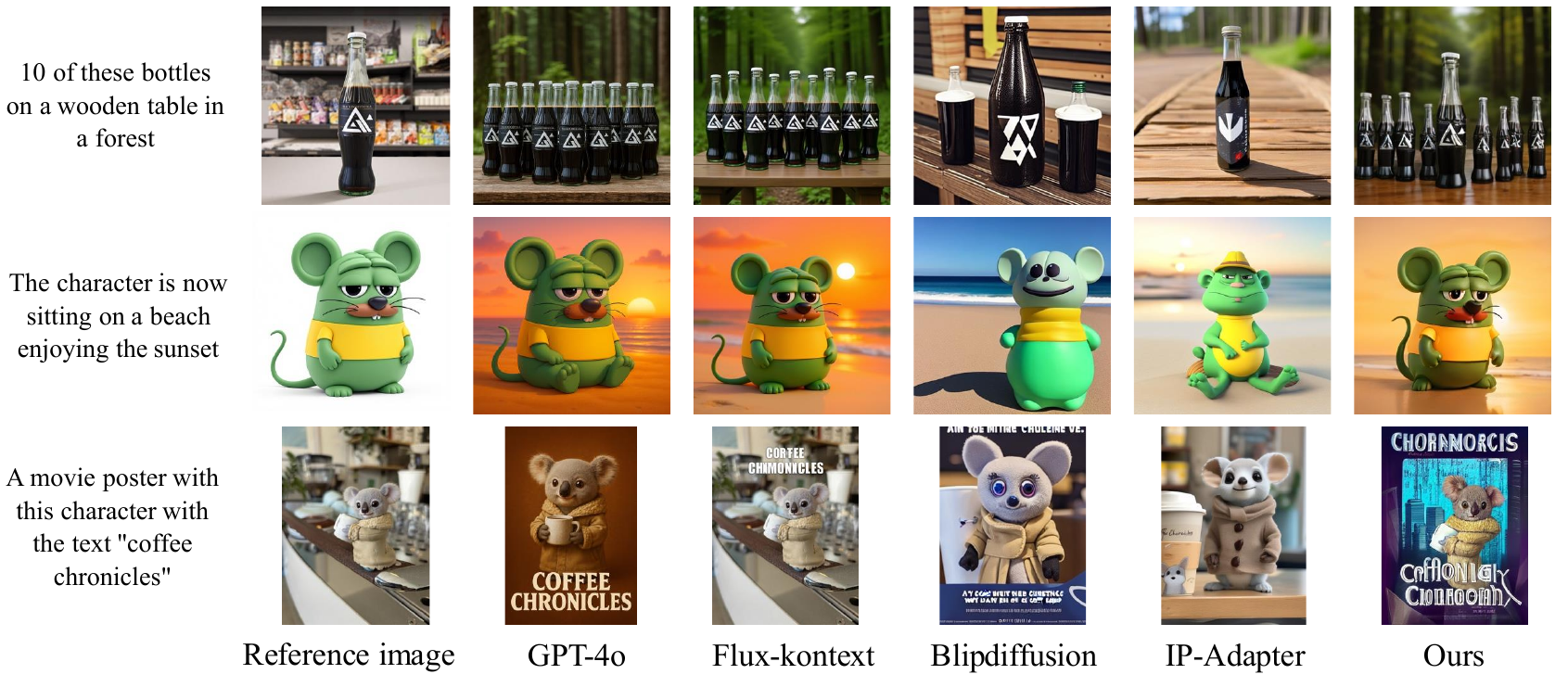}
   \caption{\textbf{Visualization of the qualitative results on Kontext-Bench.} Since this benchmark does not support subject-specific fine-tuning, we compare only with zero-shot methods on it.
   }
   \label{fig:duibijieguo_kontextimage}
\end{figure*}

\subsection{Reinforcement Learning}

The model has already acquired a certain ability to generate personalized images after supervised fine-tuning. To further enhance the  reasoning and generation capabilities of the models and ensure the format of the reasoning contents and the quality of the generation results, we treat the personalized generation task as an RL problem and adopt the GRPO strategy \cite{deepseekai2025deepseekr1incentivizingreasoningcapability} to optimize the generated images. Unlike RL methods such as PPO \cite{schulman2017proximal} that require value networks or DPO with paired preference data \cite{rafailov2023direct}, GRPO does not require explicit reward functions or human-labeled preferences. It utilizes intra-group ranking feedback to optimize the generation strategy by encouraging better samples within each group, effectively reducing computational consumption and resource requirements without the need to train additional evaluation models.

Tailored specifically for personalized generation tasks, we introduce the multi-reward mechanism and design multiple rewards which is used to enhance the consistency between the generated results and reference informations to better guide policy learning. These reward signals reflect the generation effect through groupwise preference sorting in GRPO rather than being directly used in scalar form.

\textbf{Format Reward}. To ensure that the reasoning contents generated by the model have a parsable format, we introduce the Format Reward \cite{ouyang2025motion} $R_{format}$.  This reward detects whether the result generated strictly adheres to the predefined format through a regularized expression: \verb+{<text><|image|><text>}<|image|>+. Here, the curly braces and the contents they contain represent the reasoning content, Specifically, the generated text thinking content and the corresponding image tokens that are used to help the model to understand the subject and to generate. The remaining contents are the generated result.

\textbf{Text Alignment Reward.} To ensure that the generated images are consistent with the input prompt in semantic content, we use the PickScore \cite{kirstain2023pick} metric as the standard to evaluate the quality of the generated results. It mimics the preference judgments of human users, directly optimizes the model to predict the images that users prefer, and scores them based on text conditions, making it more suitable for evaluating and optimizing the semantic consistency and user satisfaction of text-generated images. We calculate the score between the input text prompt and the generated result as $R_{t}$.

\textbf{Subject Similarity Reward}. To enable the model to learn the ability to generate personalized images from the original image to the target image, we use DreamSim \cite{fu2023dreamsim} to calculate the similarity between the reference image and the result generated as $R_{i}$. It is a metric for evaluating visual similarity based on synthetic image triples and human judgment, focusing on high-level features such as foreground and semantic information, while also taking into account low-level features like color and layout.

\section{Experiments}
\subsection{Implementation Details}
We construct the X-CoT dataset using Qwen2.5-VL-7B-Instruct \cite{bai2025qwen2} and Flux-Kontext \cite{labs2025flux}, which generate step-by-step cross-modal reasoning contents from user-provided images and prompts. These contents serve as semantic guidance for personalized image generation. We adopt Lumina-mGPT \cite{liu2024lumina} as the unified backbone model and train it in two stages: 16K steps of cold-start training on the constructed X-CoT dataset, followed by 500 steps of GRPO-based reinforcement learning using subject fidelity and text consistency as reward signals. The training process is carried out on NVIDIA A6000 GPUs. All tests are conducted on NVIDIA A100 GPUs. More experimental settings and details are provided in the supplementary materials.

\subsection{Experimental Settings}
\subsubsection{Evaluation Dataset.} We evaluate our method on two benchmarks: DreamBench \cite{ruiz2023dreambooth} and Kontext-Bench \cite{labs2025flux}. DreamBench consists of 30 subjects, each with 4–6 reference images and 25 prompts covering diverse scenes and subject variations. The subject categories span pets, household objects, and artistic sculptures, providing a broad testbed for evaluating personalization and content controllability. For Kontext-Bench, we focus on the Character Reference subset and sample approximately 200 test cases, each containing one reference image and a prompt requiring identity-consistent image generation in novel contexts. For the results of the zero-shot methods on DreamBench, we generated each image sample in the same category and took the average.

\subsubsection{Evaluation Metrics.} Following standard protocols for personalized generation \cite{ruiz2023dreambooth}, we adopt DINO~\cite{caron2021emerging} and CLIP-I \cite{radford2021learning} to assess subject fidelity, and CLIP-T \cite{radford2021learning} to assess the consistency between generated images and input text prompts. These metrics embed images and text into a shared semantic space and compute similarity scores to reflect subject fidelity and text-image alignment. For each subject, we average the scores across all prompts to obtain the final evaluation results. The specific models are ViT-B/32 and facebook/dinov2-large \cite{oquab2024dinov2learningrobustvisual} respectively.

\begin{table*}[t]
    \centering
    \setlength{\tabcolsep}{0.6mm}
    \renewcommand{\arraystretch}{1.3}
    \begin{adjustbox}{width=0.8\textwidth, center}
        \begin{tabular}{c|c||c|c|c||c|c|c||c}
            \toprule
            \multirow{2}{*}{\textbf{Type}} & \multirow{2}{*}{\textbf{Methods}} &
            \multicolumn{3}{c||}{\textbf{DreamBench}} & 
            \multicolumn{3}{c||}{\textbf{Kontext-Bench}} & 
            \multirow{2}{*}{\textbf{Zero-Shot}} \\
            \cmidrule(lr){3-8}
            & & \textbf{DINO}$\uparrow$ & \textbf{CLIP-I}$\uparrow$ & \textbf{CLIP-T}$\uparrow$ & 
            \textbf{DINO}$\uparrow$ & \textbf{CLIP-I}$\uparrow$ & \textbf{CLIP-T}$\uparrow$ & \\
            \midrule
            \multirow{4}{*}{Diffusion Model}
            & Dreambooth & 0.631 & 0.803 & 0.305 & - & - & - & No \\
            & Blip-Diffusion & 0.660 & 0.818 & 0.283 & 0.500 & 0.742 & 0.246 & Yes \\
            & IP-Adapter & 0.671 & 0.836 & 0.291 & 0.516 & \textbf{0.762} & 0.249 & Yes \\
            & Flux-Kontext & 0.682 & \textbf{0.846} & \underline{0.310} & \underline{0.554} & \underline{0.750} & \textbf{0.308} & Yes \\
            \midrule
            \multirow{4}{*}{Unified MLLM}
            & Yo'Chameleon & 0.542 & 0.795 & 0.225 & - & - & - & No \\
            & UniCTokens & 0.599 & 0.782 & 0.304 & - & - & - & No  \\
            & GPT-4o & \underline{0.722} & 0.803 & 0.274 & 0.501 & 0.725 & \underline{0.303} & Yes  \\
            & Ours & \textbf{0.786} & \underline{0.842} & \textbf{0.313} & \textbf{0.562} & \underline{0.750} & 0.296 & Yes \\
            \bottomrule
        \end{tabular}
    \end{adjustbox}
    \caption{\textbf{Quantitative comparison on DreamBench and Kontext-Bench.} Our MM-R1 achieves the best results across multiple metrics, notably scoring highest on DreamBench (DINO: 0.786, CLIP-T: 0.313) and Kontext-Bench (DINO: 0.562), while also supporting zero-shot capabilities. For each test sample we conduct four tests and take the average as the result.
    }
    \label{tab:zhubiao}
\end{table*}

\begin{figure}[t]
  \centering
  \includegraphics[width=\linewidth]{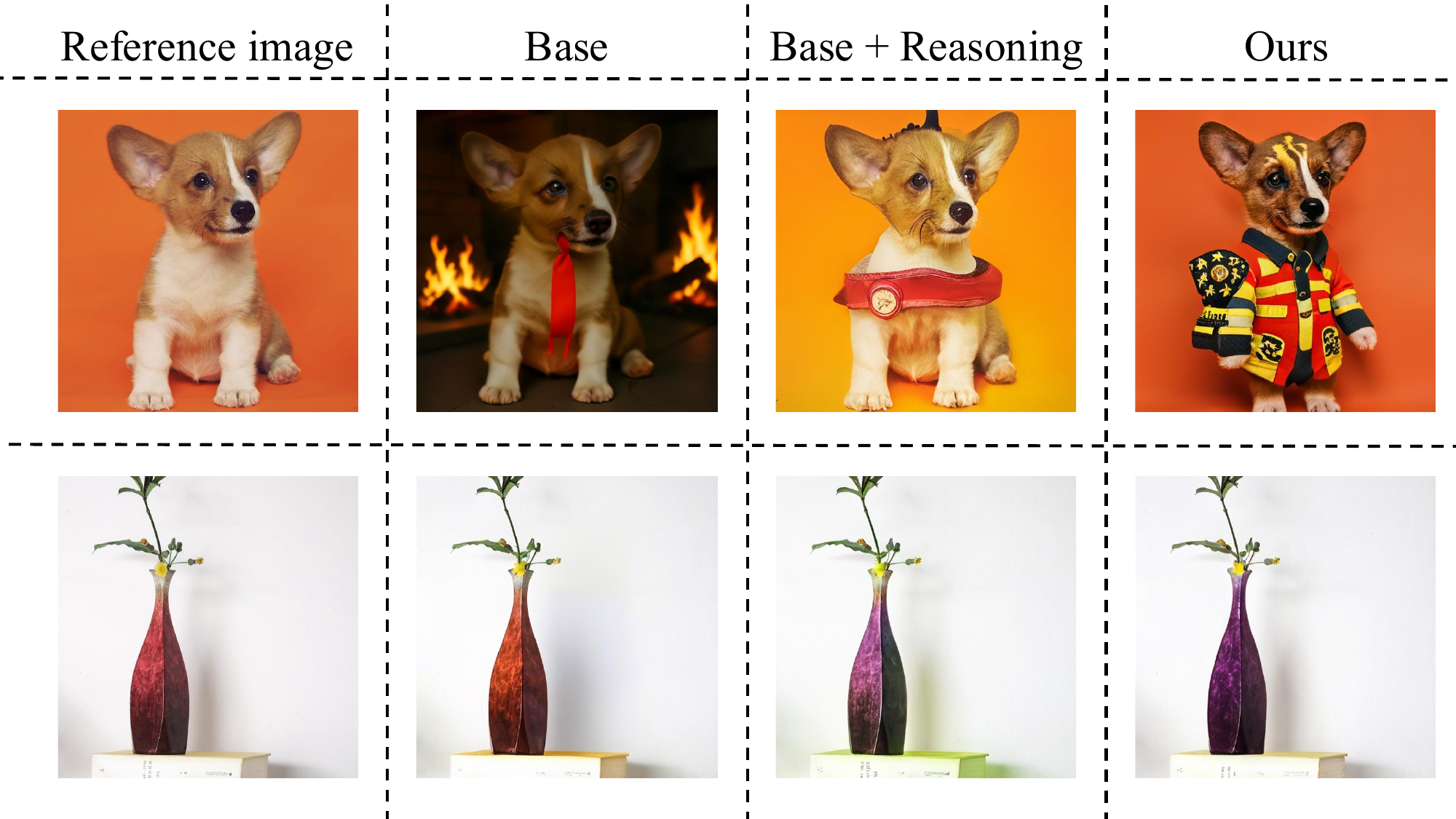}
  \caption{\textbf{Comparison of results in the ablation experiment.} \textit{Base} means that two images are generated using the base model, and \textit{Base + Reasoning} represents generating images using the Base model and reasoning generation strategy. Prompt is \textit{``a dog in a fireman outfit.''}.}
  \label{fig:xianrongjieguo_image}
\end{figure}

\subsection{Quantitative Evaluation}
We evaluate the performance of MM-R1 on the DreamBench and Kontext-Bench, comparing it with state-of-the-art methods across both diffusion-based and unified MLLM-based approaches (Tab. \ref{tab:zhubiao}). For diffusion models, we include DreamBooth \cite{ruiz2023dreambooth}, Blip-Diffusion \cite{li2023blip}, IP-Adapter \cite{ye2023ip}, and Flux-Kontext~\cite{labs2025flux}. For unified MLLMs, we compare against Yo'Chameleon \cite{nguyen2025yo}, UniCTokens \cite{an2025unictokens}, and GPT-4o \cite{hurst2024gpt}. Among them, although Yo'Chameleon and UniCTokens can perform both understanding and generation tasks, here we only test the generation task and do not focus on their performance in the understanding task.

On DreamBench, MM-R1 achieves the best DINO score (0.786) and CLIP-T score (0.313), indicating strong subject fidelity and prompt alignment. On Kontext-Bench, it also ranks first in DINO (0.562) and performs competitively in CLIP-I (0.750) and CLIP-T (0.296), validating its effectiveness in single-image reference generation. Overall, MM-R1 establishes a new state-of-the-art among unified MLLMs and achieves performance that is competitive with the strongest diffusion-based approaches, all without relying on subject-specific fine-tuning.

\subsection{Qualitative Results}
Fig. \ref{fig:duibijieguo_dbimage} and \ref{fig:duibijieguo_kontextimage} show the comparison of qualitative results between our method and some other methods. It can be seen that our method can maintain a high level of image fidelity and text controllability. 
As shown in our picture (the last row of the third column), the rc\_car is highly similar not only overall but also in detail to the rc\_car in the reference image. This indicates that, through X-CoT, the model can extract more attributes of the reference subject to generate consistent images.
Furthermore, our method can generate more diverse images, such as the last column of the second row in Fig. \ref{fig:duibijieguo_dbimage}. This is because our method performs information integration and subject extraction before generating images, which can help the model decouple subject from input and reduce the interference of irrelevant information. 
More test sample images and human evaluation results will appear in the supplementary materials.

\begin{table}[t]
\centering
\setlength{\tabcolsep}{2pt}
\resizebox{\columnwidth}{!}{
  \begin{tabular}{lccc|ccc}
    \toprule
    \multirow{2}{*}{Method} 
    & \multicolumn{3}{c|}{\textbf{DreamBench}} 
    & \multicolumn{3}{c}{\textbf{Kontext-Bench}} \\
    \cmidrule(lr){2-4} \cmidrule(lr){5-7}
    & DINO $\uparrow$ & CLIP-I $\uparrow$ & CLIP-T $\uparrow$ 
    & DINO $\uparrow$ & CLIP-I $\uparrow$ & CLIP-T $\uparrow$ \\
    \midrule
    $R_{f}$ & 0.729 & 0.768 & 0.298 & 0.512 & 0.712 & 0.284 \\ 
    $R_{f} + R_{i}$ & \underline{0.782} & \textbf{0.845} & 0.284 & \underline{0.559} & \underline{0.749} & 0.269 \\
    $R_{f} + R_{t}$ & 0.737 & 0.801 & \underline{0.311} & 0.525 & 0.714 & \textbf{0.301} \\
    \midrule
    \textbf{Ours} & \textbf{0.786} & \underline{0.842} & \textbf{0.313} & \textbf{0.562} & \textbf{0.750} & \underline{0.296} \\
    \bottomrule
  \end{tabular}
}
\caption{\textbf{Ablation results of different reward functions in GRPO.} Each line represents the result of GRPO reinforcement learning using different reward functions after the model has undergone X-CoT cold-start. Among them, $R_{f}$ represents Format Reward, $R_{i}$ represents Subject Similarity Reward, $R_{t}$ represents Text Alignment Reward.}
\label{tab:rlxiaorong}
\end{table}

\begin{table}[t]
\centering
\setlength{\tabcolsep}{2pt}
\resizebox{\columnwidth}{!}{
  \begin{tabular}{lccc|ccc}
    \toprule
    \multirow{2}{*}{Method} 
    & \multicolumn{3}{c|}{\textbf{DreamBench}}
    & \multicolumn{3}{c}{\textbf{Kontext-Bench}} \\
    \cmidrule(lr){2-4} \cmidrule(lr){5-7}
    & DINO $\uparrow$ & CLIP-I $\uparrow$ & CLIP-T $\uparrow$ 
    & DINO $\uparrow$ & CLIP-I $\uparrow$ & CLIP-T $\uparrow$ \\
    \midrule
    Base & 0.631 & 0.728 & 0.270 & 0.372 & 0.663 & 0.216 \\
    Reasoning & 0.650 & 0.742 & 0.278 & 0.380 & 0.683 & 0.221 \\
    w/o GRPO & 0.727 & 0.761 & \underline{0.297} & 0.505 & 0.702 & \underline{0.287} \\
    w/o X-CoT & \underline{0.732} & \underline{0.801} & 0.288 & \underline{0.525} & \underline{0.714} & 0.282 \\
    \midrule
    \textbf{Ours} & \textbf{0.786} & \textbf{0.842} & \textbf{0.313} & \textbf{0.562} & \textbf{0.750} & \textbf{0.296} \\
    \bottomrule
  \end{tabular}
}
\caption{\textbf{Ablation results of our proposed components.} The quantitative results of each behavioral model after different steps of training. Among them, \textit{Base} refers to the generation using only Lumina-mGPT, \textit{Reasoning} indicates that the base model is used for reasoning understanding first and then for generation, \textit{w/o GRPO} indicates that the model has been trained by our X-CoT, and \textit{w/o X-CoT} indicates that the base model adjusted by the GRPO.}
\label{tab:xiaorong}
\end{table}

\subsection{Ablation Study}
\subsubsection{Ablation Study on MM-R1 Modules.} To evaluate the contribution of each component in our framework, we conduct an ablation study as shown in Tab. \ref{tab:xiaorong}. We begin by evaluating our reasoning generation strategy on the base model without any additional training. In this setting, the understanding stage guides the model to interpret the reference image and generate an intermediate image that isolates the subject and corresponding text thinking contents. This result is then provided to the generation stage, together with the user prompt, to guide the final personalized image generation. As shown in Tab. \ref{tab:xiaorong} and Fig. \ref{fig:xianrongjieguo_image}, this structurally guided inference pipeline already yields noticeable improvements, highlighting the effectiveness of our X-CoT formulation.

We further evaluate the impact of supervised X-CoT training without applying GRPO. This variant enhances the model’s ability to extract subject-specific concepts and align them with user prompts, leading to clear improvements in fidelity and consistency. Conversely, we assess the effect of applying GRPO reinforcement learning without X-CoT supervision, which also results in notable gains, particularly in text alignment. When all components are combined, the model achieves the highest scores across all evaluation metrics, confirming the complementary strengths of structured reasoning and reward-based optimization.

\subsubsection{Ablation Study on Reward Design.} To evaluate the contribution of each reward component in GRPO, we conduct an ablation study as summarized in Tab. \ref{tab:rlxiaorong}. The Format reward ($R_f$) is used by default to enforce structural constraints in the X-CoT process, and also brings slight improvements in output stability. Adding the Subject Similarity Reward ($R_i$) yields notable gains in identity preservation, as it directly encourages alignment between the generated and reference images. In contrast, the Text Alignment Reward ($R_t$) enhances semantic consistency with the input prompt. When both $R_i$ and $R_t$ are incorporated, the model achieves further improvements, indicating that these two rewards offer complementary supervision signals during training.

\section{Conclusion}
In this paper, we introduce MM-R1 which achieving personalization by directly enhancing the MLLMs' intrinsic reasoning abilities. Inspired by the relevance of understanding and generating tasks in MLLMs, we divide the personalization task into understanding the attributes of subject and image generation, and using this strategy to propose the X-CoT framework for image generation. To better unleash the model's capabilities of understanding and generation, we train the model that has been fine-tuned with the X-CoT strategy using GRPO. The experiments show that this strategy can effectively help the model better complete the personalized generation task.

\bibliography{aaai2026lq}

\end{document}